\documentclass[11pt,a4paper]{article}
\usepackage[hyperref]{emnlp2018}
\usepackage{times}
\usepackage{latexsym}
\usepackage{graphicx}
\usepackage{url}
\aclfinalcopy % Uncomment this line for the final submission
\usepackage{flushend}

\title{Evaluating the Underlying Gender Bias in \\Contextualized Word Embeddings}

\author{Christine Basta\hspace{7mm} Marta R. Costa-juss\`a \hspace{7mm} Noe Casas\\ 
  Universitat Polit\`ecnica de Catalunya\\
  \texttt{\{christine.raouf.saad.basta,marta.ruiz,noe.casas\}@upc.edu} \\
  \\}
\date{}

\begin{document}
\maketitle
\begin{abstract}
Gender bias is highly impacting natural language processing applications. Word embeddings have clearly been proven both to keep and amplify gender biases that are present in current data sources. Recently, contextualized word embeddings have enhanced previous word embedding techniques by computing word vector representations dependent on the sentence they appear in. 

In this paper, we study the impact of this conceptual change in the word embedding computation in relation with gender bias. Our analysis includes different measures previously applied in the literature to standard word embeddings. Our
findings suggest that contextualized word embeddings are less biased than
standard ones even when the latter are debiased.
\end{abstract}

\section{Introduction}
Social biases in machine learning, in general and in natural language processing (NLP) applications in particular, are raising the alarm of the scientific community. Examples of these biases are evidences such that face recognition systems or speech recognition systems works better for white men than for ethnic minorities \cite{DBLP:conf/fat/BuolamwiniG18}. Examples in the area of NLP are the case of machine translation that systems tend to ignore the coreference information in benefit of an stereotype \cite{font:2019} or sentiment analysis where higher sentiment intensity prediction is biased for a particular gender \cite{kiritchenko:2018}. 

%NLP models are trained on human language datasets, which can represent a significant part of the problem due to the existence of bias in it and due to the under representation of women in such data \cite{leavy:2018}. Beyond detection, there are already other works that are proposing debiasing methods, most of which are related to data augmentation \cite{madaan:2018,Lu:2018}. 

%For example, researchers in \cite{madaan:2018,Lu:2018} targeted debiasing the data at the source by augmenting or varying data with counter based examples of gender pronouns for sentences, which have a mentioned profession.  Another \newcite{Zhao2017} propose interventions and constraints at the corpus level. Other more complex methods propose particular techniques to debias algorithms themselves and have been tested for example in language modeling \cite{}, sentiment analysis \cite{} and machine translation \cite{}.

In this work we focus on the 
particular NLP area of word embeddings \cite{mikolov:2010}, which
represent words in a numerical vector space. Word embeddings
representation spaces are known to present geometrical phenomena
mimicking relations and analogies between words (e.g. \textit{man} is to \textit{woman} as \textit{king} is to \textit{queen}). Following this property of finding relations or analogies, one popular example of gender bias is the word association between \textit{man} to  \textit{computer programmer} as \textit{woman} to \textit{homemaker} \cite{bolukbasi:2016}. Pre-trained word embeddings are used in many NLP downstream tasks, such as natural language inference (NLI), machine translation (MT) or question answering (QA). Recent progress in word embedding techniques has been achieved with contextualized word embeddings \cite{elmo} which provide different vector representations for the same word in different contexts.

While gender bias has been studied, detected and partially addressed for standard word embeddings techniques \cite{bolukbasi:2016,zhao:2018,lipstick:2019}, it is not the case for the latest techniques of contextualized word embeddings. Only just recently, \newcite{zhao:2019} present a first analysis on the topic based on the proposed methods in \newcite{bolukbasi:2016}. In this paper, we further analyse the presence of gender biases in contextualized word embeddings by means of the proposed methods in \newcite{lipstick:2019}. For this, in section \ref{sec:background} we
provide an overview of the relevant work on which we build our analysis;
in section \ref{sec:questions} we state the specific request questions
addressed in this work, while in section \ref{sec:expframework} we describe the
experimental framework proposed to address them and in section
\ref{sec:evaluation} we present the obtained and discuss the results; finally, in
section \ref{sec:conclusions} we draw the conclusions of our work and propose some further research.

%\section{Related work}
%\label{sec:relwork}

%Gender bias has gained much attention in the recent years, due to its high impact on natural language processing applications.  

%{\color{red} General techniques of sentimental, MT, Coreference, language model, small hint for embeddinsg}

\section{Background}
\label{sec:background}

In this section we describe the relevant NLP techniques used
along the paper, including word embeddings, their
debiased version and contextualized word representations.

\subsection{Words Embeddings}

Word embeddings are distributed representations in a vector space. These
vectors are normally learned from large corpora and are then used in
downstream tasks like NLI, MT, etc. Several approaches have been proposed
to compute those vector representations, with word2vec \cite{mikolov:2013}
being one of the dominant options. Word2vec proposes two variants:
continuous bag of words (CBoW) and skipgram, both consisting of a single hidden
layer neural network train on predicting a target word from its context words
for CBoW, and the opposite for the skipgram variant.
The outcome of word2vec is an embedding table, where a numeric vector
is associated to each of the words included in the vocabulary.

These vector representations, which in the end are computed on
co-occurrence statistics, exhibit geometric properties resembling
the semantics of the relations between words. This way, subtracting
the vector representations of two related words and adding the
result to a third word, results in a representation that
is close to the application of the semantic relationship between the two
first words to the third one. This application of analogical relationships have been
used to showcase the bias present in word embeddings, with the prototypical
example that when subtracting the vector representation of \textit{man} from
that of \textit{computer} and adding it to \textit{woman}, we obtain
\textit{homemaker}.

\subsection{Debiased Word Embeddings}

Human-generated corpora suffer from social biases. Those biases are reflected
in the co-occurrence statistics, and therefore learned into word embeddings
trained in those corpora, amplifying them
\cite{bolukbasi:2016,caliskan:2017}.

\newcite{bolukbasi:2016} studied from a geometrical point of view
the presence of gender bias in word embeddings. For this, they
compute the subspace where the gender information concentrates
by computing the principal components of the difference of
vector representations of male and female gender-defining word pairs.
With the gender subspace, the authors identify 
direct and indirect biases in profession words. Finally,
they mitigate the bias by nullifying the information in the
gender subspace for words that should not be associated to gender,
and also equalize their distance to both elements of gender-defining
word pairs.

\newcite{zhao2018learning} proposed an extension to GloVe embeddings
\cite{pennington:2014} where the loss function used to train the
embeddings is enriched with terms that confine the gender
information to a specific portion of the embedded vector. The
authors refer to these pieces of information as \textit{protected attributes}.
Once the embeddings are trained, the gender protected attribute can
be simply removed from the vector representation, therefore eliminating any
gender bias present in it.

The transformations proposed by both \newcite{bolukbasi:2016} and
\newcite{zhao2018learning} are downstream task-agnostic. This fact is
used in the work of \newcite{lipstick:2019} to showcase that, while
apparently the embedding information is removed, there is still gender
information remaining in the vector representations.

\subsection{Contextualized Word Embeddings}

Pretrained Language Models (LM) like
ULMfit \cite{howard2018ulmfit},
ELMo \cite{elmo},
OpenAI GPT \cite{radford2018gpt1,radford2019gpt2}
and BERT \cite{bert}, proposed different
neural language model architectures and made their pre-trained
weights available to ease the application of transfer learning
to downstream tasks, 
where they have pushed the state-of-the-art for
several benchmarks including question answering on SQuAD,
NLI, cross-lingual NLI and named identity recognition (NER).

While some of these pre-trained LMs,
like BERT, use subword level tokens, ELMo provides
word-level representations. \newcite{peters2019iceandfireemoji} and 
\newcite{liu2019transferability} confirmed the viability of using
ELMo representations directly as features for downstream tasks without 
re-training the full model on the target task.

Unlike word2vec vector representations, which are constant
regardless of their context, ELMo representations depend
on the sentence where the word appears, and therefore the full model
has to be fed with each whole sentence to get the word representations.

The neural architecture proposed in ELMo \cite{elmo} consists of a 
character-level convolutional layer processing the characters
of each word and creating a word representation that is then fed to
a 2-layer bidirectional LSTM \cite{hochreiter1997lstm},
trained on language modeling task on a large corpus.

\section{Research questions}
\label{sec:questions}

Given the high impact of contextualized word embeddings in the area of NLP and the social consequences of having biases in such embeddings, in this work we
analyse the presence of bias in these contextualized word embeddings.
In particular, we focus on gender biases, and specifically on the following questions:
\begin{itemize}
\item Do contextualized word embeddings exhibit gender bias and how does this bias compare to standard and debiased word embeddings?

%\item Can the same techniques of quantifying bias in standard embeddings be applied on contextualized embeddings?

\item  Do different evaluation techniques identify bias similarly and what would be the best measure to use for gender bias detection in contextualized embeddings?
\end{itemize}

To address these questions, we adapt and contrast with the evaluation measures proposed by \newcite{bolukbasi:2016} and \newcite{lipstick:2019}.

\section{Experimental Framework}
\label{sec:expframework}

%We evaluated the bias in contextualized embeddings inspired from both Bolukbasi \cite{bolukbasi:2016} and Hila \cite{lipstick:2019}. 

%DONE! {\color{red} TODO:citations to corpora}

As follows, we define the data and resources that we are using for performing our experiments. We also motivate the approach that we are using for contextualized word embeddings.

We worked with the English-German news corpus from the WMT18\footnote{\url{http://data.statmt.org/wmt18/translation-task/training-parallel-nc-v13.tgz}}. We used the English side with 464,947 lines and 1,004,6125 tokens. %We prepared the swapped version for this file, as illustrated in the first experiment.

To perform our analysis we used a set of lists from previous work \cite{bolukbasi:2016,lipstick:2019}. We refer to the list of definitional pairs \footnote{\url{https://github.com/tolga-b/debiaswe/blob/master/data/definitional\_pairs.json}} as  'Definitonal List' (e.g. \textit{she}-\textit{he}, \textit{girl}-\textit{boy}). We refer to the list of female and male professions \footnote{\url{https://github.com/tolga-b/debiaswe/blob/master/data/professions.json}} as 'Professional List' (e.g. \textit{accountant}, \textit{surgeon}). The 'Biased List' is the list used in the clustering experiment and it consists of biased male and female words (500 female biased tokens and 500 male biased token). 
This list is generated by taking the most biased words, where the bias of a word is computed by taking its projection on the gender direction (\overrightarrow{he}-\overrightarrow{she}) (e.g. \textit{breastfeeding}, \textit{bridal} and \textit{diet} for female and \textit{hero}, \textit{cigar} and \textit{teammates} for male). The 'Extended Biased List' is the list used in classification experiment
, which contains 5000 male and female biased tokens, 2500 for each gender, generated in the same way of the Biased List\footnote{Both 'Biased List' and 'Extended Biased List' were kindly provided by Hila Gonen to reproduce experiments from her study \cite{lipstick:2019}}.
A note to be considered, is that the lists we used in our experiments (and obtained from \newcite{bolukbasi:2016} and \newcite{lipstick:2019}) may contain words that are missing in our corpus and so we can not obtain contextualized embeddings for them.

Among different approaches to contextualized word embeddings (mentioned in section \ref{sec:background}), we choose 
%the motivation for choosing
ELMo \cite{elmo} as contextualized word embedding
approach. The motivation for using ELMo instead of other approaches like BERT \cite{bert} is that
ELMo provides word-level representations, as opposed to BERT's subwords.
This makes it possible to study the word-level semantic traits directly,
without resorting to extra steps to compose word-level information from the
subwords that could interfere with our analyses.

\section{Evaluation measures and results}
\label{sec:evaluation}

There is no standard measure for gender bias, and even less for such
the recently proposed contextualized word embeddings.
In this section, we adapt gender bias measures for word embedding methods
from previous work \cite{bolukbasi:2016} and \cite{lipstick:2019} to
be applicable to contextualized word embeddings.

This way, we first compute the gender subspace from the ELMo
vector representations of gender-defining words, then
identify the presence of direct bias in the contextualized
representations. We then proceed to
identify gender information by means of clustering and
classifications techniques. We compare our results to previous results from debiased and non-debiased word embeddings \cite{bolukbasi:2016} .

\paragraph{Detecting the Gender Space}\label{para_detecting_gender}

\begin{figure*}
 \center
  \includegraphics[width=\textwidth]{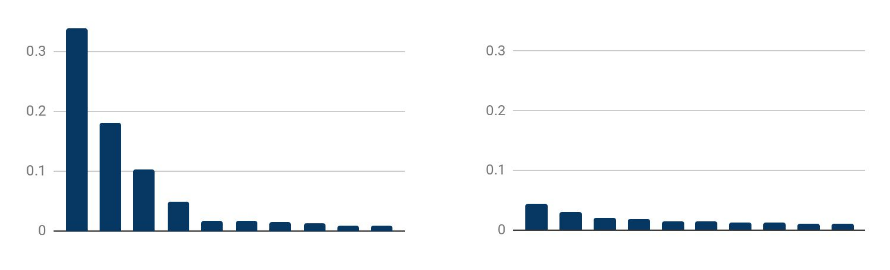}
  \caption{(Left) the percentage of variance explained in the PC of definitional vector differences. (Right) The corresponding percentages for random vectors.}
  \label{fig:gender_bias_figure}
\end{figure*}

\newcite{bolukbasi:2016} propose to identify gender bias in word representations
by computing the direction between representations of male and female
word pairs from the Definitional List (\overrightarrow{he}-\overrightarrow{she}, \overrightarrow{man}-\overrightarrow{woman}) and computing their principal components.

In the case of contextualized embeddings, there is not just a single
representation for each word, but its representation depends on the
sentence it appears in. This way, in order to compute the gender subspace
we take the representation of words by randomly sampling sentences that
contain words from the Definitional List and,
for each of them, we swap the definitional word with its pair-wise equivalent
from the opposite gender.
We then obtain the ELMo representation of the definintional word in
each sentence pair, computing their difference.
On the set of difference vectors, we compute their principal components
to verify the presence of bias. In order to have a reference, we computed
the principal components of representation of random words.

Similarly to \newcite{bolukbasi:2016}, figure \ref{fig:gender_bias_figure}
shows that the first eigenvalue is significantly larger than the rest and
that there is also a single direction describing the majority of variance
in these vectors, still the difference between the percentage of variances
is less in case of contextualized embeddings, which may refer that there
is less bias in such embeddings. We can easily note the difference in the
case of random, where there is a smooth and gradual decrease in eigenvalues,
and hence the variance percentage.

A similar conclusion was stated in the recent work \cite{zhao:2019} where the
authors applied the same approach, but for gender swapped variants of sentences
with professions. They computed the difference between the vectors of occupation
words in corresponding sentences and got a skewed graph where the first component
represent the gender information while the second component groups the male and
female related words. 

\paragraph{Direct Bias}

Direct Bias is a measure of how close a certain set of words are to the gender
vector. To compute it, we extracted from the training data the sentences that contain words in the Professional List. We excluded the sentences that have both a professional token and definitional gender word to avoid the influence of the latter over the presence of bias in the former. We applied the definition of direct bias from \newcite{bolukbasi:2016} on the ELMo representations of the professional words in these sentences.
\begin{equation}
\frac{1}{|N|}\sum_{w \epsilon N}|cos(\vec{w},g)|
\end{equation}

where N is the amount of gender neutral words, $g$ the gender direction, and $\vec{w}$ the word vector of each profession. We got direct bias of 0.03, compared to 0.08 from standard word2vec embeddings described in \newcite{bolukbasi:2016}. This reduction on the direct bias confirms that the substantial component along the gender direction that is present in standard word embeddings is less for the contextualized word embeddings. Probably, this reduction comes from the fact that we are using different word embeddings for the same profession depending on the sentence which is a direct consequence and advantage of using contextualized embeddings.

\paragraph{Male and female-biased words clustering.}

In order to study if biased male and female words cluster together when applying contextualized embeddings, we used k-means to generate 2 clusters of the embeddings of tokens from the Biased list. 
Note that we can not use several representations for each word, since it would not make any sense to cluster one word as male and female at the same time. Therefore, in order to make use of the advantages of the contextualized embeddings, we repeated 10 independent experiments, each with a different random sentence of each word from the list of biased male and female words.

Among these 10 experiments, we got a minimum accuracy of 69.1\% and a maximum of 71.3\%, with average accuracy of 70.1\%, much lower than in the case of biased and debiased word embeddings which were 99.9 and 92.5, respectively, as stated in \newcite{lipstick:2019}. Based on this criterion, even if there is still bias information to be removed from contextualized embeddings, it is much less than in case of standard word embeddings, even if debiased.  

\begin{figure}
 \center
  \includegraphics[width=0.5\textwidth]{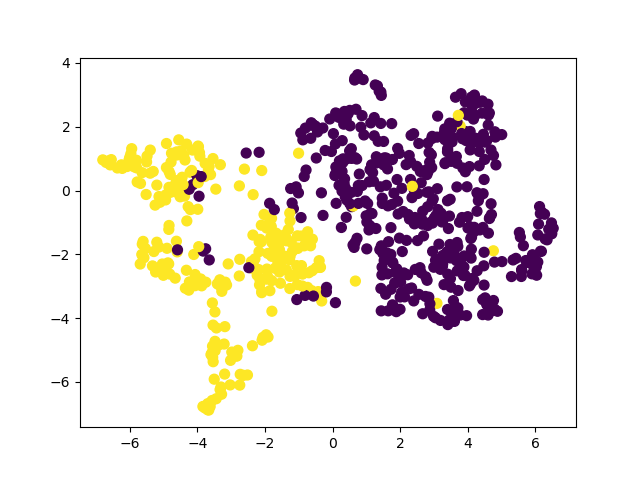}
  \caption{K-means clustering, the yellow color represents
  the female and the violet represents the male}
  \label{fig:clustering_500_figure}
\end{figure}

The clusters (for one particular experiment out of the 10 of them) are shown in Figure \ref{fig:clustering_500_figure} after applying UMAP \cite{umap2018algo,umap2018software} to the contextualized embeddings. 

\paragraph{Classification Approach}

In order to study if contextualized embeddings learn to generalize bias, we trained a Radial Basis Function-kernel Support Vector Machine classifier on the embeddings of random 1000 biased words from the Extended Biased List. After that, we evaluated the generalization on the other random 4000 biased tokens. Again, we performed 10 independent experiments, to guarantee randomization of word representations. Among these 10 experiments, we got a minimum accuracy of 83.33\% and a maximum of 88.43\%, with average accuracy of 85.56\%. This number shows that the bias is learned in these embeddings with high rate. However, it learns in less rate than the normal embeddings, whose classification reached 88.88\% and 98.25\% for biased and debiased versions, respectively.

\paragraph{K-Nearest Neighbor Approach}
\label{sec:experiments}

\begin{figure}
 \center
  \includegraphics[width=0.5\textwidth]{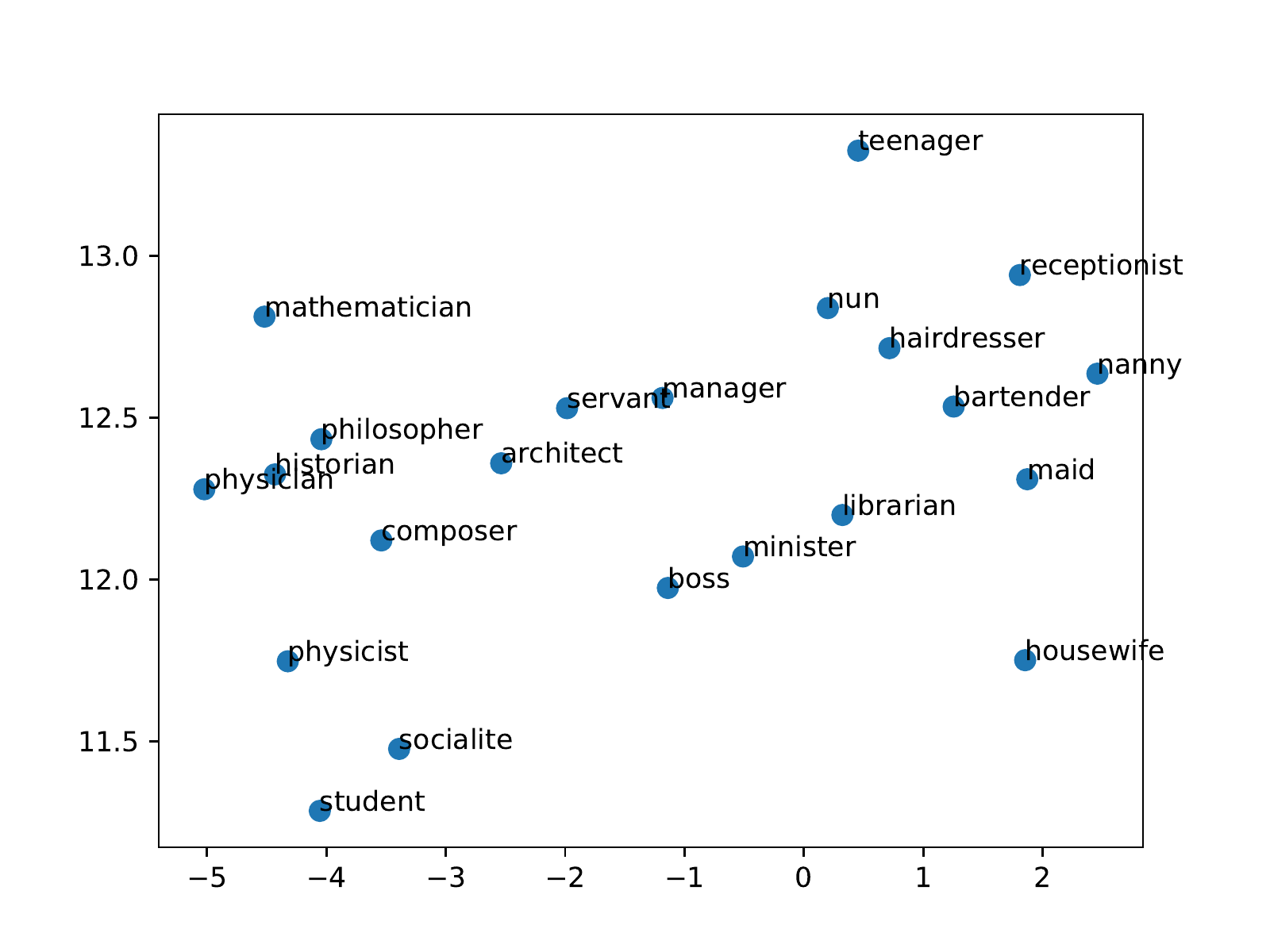}
  \caption{Visualization of contextualized embeddings of professions.}
  \label{fig:knn_professions}
\end{figure}

To understand more about the bias in contextualized embeddings, it is important to analyze the bias in the professions. The question is whether these embeddings stereotype the professions as the normal embeddings. This can be shown by the nearest neighbors of the female and male stereotyped professions, for example 'receptionist' and 'librarian' for female and 'architect' and 'philosopher' for male.  We applied the k nearest neighbors on the Professional List, to get the nearest k neighbor to each profession. We used a random representation for each token of the profession list, after applying the k nearest neighbor algorithm on each profession, we computed the percentage of female and male stereotyped professions among the k nearest neighbor of each profession token. Afterwards, we computed the Pearson correlation of this percentage with the original bias of each profession. Once again, to assure randomization of tokens, we performed 10 experiments, each with different random sentences for each profession, therefore with different word representations. The minimum Pearson correlation is 0.801 and the maximum is 0.961, with average of 0.89. All these correlations are significant with p-values smaller than $1 \times 10^{-40}$. This experiment showed the highest influence of bias compared to 0.606 for debiased embeddings and 0.774 for non-debiased. Figure \ref{fig:knn_professions} demonstrates this influence of bias by showing that female biased words (e.g. \textit{nanny}) has higher percent of female words than male ones and vice-versa for male biased words (e.g. \textit{philosopher}).

\section{Conclusions and further work}
\label{sec:conclusions}

While our study can not draw clear conclusions on whether contextualized word embeddings augment or reduce the gender bias, our results show more insights of which aspects of the final contextualized word vectors get affected by such phenomena, with a tendency more towards reducing the gender bias rather than the contrary. 

Contextualized word embeddings mitigate gender bias when measuring in the following aspects:

\begin{enumerate}
\item Gender space, which is capturing the gender direction from word vectors, is reduced for gender specific contextualized word vectors compared to standard word vectors. 

\item Direct bias, which is measuring how close set of words are to the gender vector, is lower for contextualized word embeddings than for standard ones.

\item Male/female clustering, which is produced between words with strong gender bias, is less strong than in debiased and non-debiased standard word embeddings.

\end{enumerate}

However, contextualized word embeddings preserve and even amplify gender bias when taking into account other aspects:

\begin{enumerate}

\item The implicit gender of words can be predicted with accuracies higher than 80\% based on contextualized word vectors which is only a slightly lower accuracy than when using vectors from debiased and non-debiased standard word embeddings.

\item The stereotyped words group with implicit-gender words of the same gender more than in the case of debiased and non-debiased standard word embeddings.

\end{enumerate}

While all measures that we present exhibit certain gender bias, when evaluating future debiasing methods for contextualized word embeddings it would be worth it putting emphasis on the latter two evaluation measures that show higher bias than the first three.

Hopefully, our analysis will provide a grain of sand towards defining standard evaluation methods for gender bias, proposing effective debiasing methods or even directly designing equitable algorithms which automatically learn to ignore biased data. 

%{\color{red} answer:which are the measures from bolukbasi and hila that we have not addressed? we should name them for further work}

As further work, we plan to extend our study to multiple domains and multiple languages to analyze and measure the impact of gender bias present in contextualized embeddings in these different scenarios.

\section*{Acknowledgements}

We want to thank Hila Gonen for her support during our research.

This work is supported in part by the Catalan Agency for Management of University and Research Grants (AGAUR) through the FI PhD Scholarship and the Industrial PhD Grant.
This work is also supported in part by the
Spanish Ministerio de Econom\'ia y Competitividad,
the European Regional Development Fund
and the Agencia Estatal de Investigaci\'on,
through the postdoctoral senior grant Ram\'on y Cajal, contract TEC2015-69266-P
(MINECO/FEDER,EU) and contract PCIN-2017-079 (AEI/MINECO). 

\bibliographystyle{acl_natbib_nourl}
\bibliography{relwork}

%\appendix

\end{document}